# EFFICIENT KNOWLEDGE BASE MANAGEMENT IN DCSP

Hong Jiang

Mathematics & Computer Science Department, Benedict College, USA

jiangh@benedict.edu

## ABSTRACT

*DCSP (Distributed Constraint Satisfaction Problem) has been a very important research area in AI (Artificial Intelligence). There are many application problems in distributed AI that can be formalized as DSCPs. With the increasing complexity and problem size of the application problems in AI, the required storage place in searching and the average searching time are increasing too. Thus, to use a limited storage place efficiently in solving DCSP becomes a very important problem, and it can help to reduce searching time as well. This paper provides an efficient knowledge base management approach based on general usage of hyper-resolution-rule in consistence algorithm. The approach minimizes the increasing of the knowledge base by eliminate sufficient constraint and false nogood. These eliminations do not change the completeness of the original knowledge base increased. The proofs are given as well. The example shows that this approach decrease both the new nogoods generated and the knowledge base greatly. Thus it decreases the required storage place and simplify the searching process.*

## 1. INTRODUCTION

Many problem solving techniques in AI (Artificial Intelligence) are searching based. And the problems that have been addressed upon search algorithms can be categorized as three classes as mentioned in [1]: (1) Constraint satisfaction problems (CSP), (2) path-finding problems, and (3) two-player games. Among them, CSP and distributed version of CSP become a very important research area, because many problems that arise in multiagent systems can be reduced to a distributed constraint satisfaction problem(DCSP) and this approach has led to many successful multiagent applications.

Hyper-resolution-rule [2, 3] is a basic unit used in many famous algorithms used in solving DCSPs, such as asynchronous backtracking algorithm(ABT) [4–8] and asynchronous weak-commitment search algorithm [9–11]. However the detailed usage of the hyper-resolution-rule in those algorithms is not clearly defined, and the general usage of the hyper-resolution-rule for consistence algorithm produces polynomial space usage. This is still fordable for some small size problems, but for bigger size problems polynomial space usage will become a big problem. Practically, most of the problems in multiagent system become bigger and bigger. Then the space usage problem can not be ignored, and research on how to efficiently use limited storage place in solving DCSP becomes important.

This paper provides an efficient knowledge base management approach based on general usage of hyper-resolution-rule in consistency algorithm [12]. This approach minimizes the increasing of the knowledge base by eliminate sufficient constraint and false nogood, which are defined in section 3.1. These eliminations do not change the completeness of the original knowledge base increased. The proofs are given in section 3.1.

In this paper, section 2.1 gives description about CSP and DCSP; section 2.2 describes the general Hyper-Resolution-Based Consistency Algorithm and the related Polynomial Space Problem; section 3.1 derives some theorems from hyper-resolution-rule, which are the foundation of our approach based on; section 3.2 describes our approach; example to compare the general approach with our approach is given in section 4; finally, evaluation and conclusion are given in section 5 and 6.

## 2. HYPER-RESOLUTION BASED POLYNOMIAL SPACE PROBLEM IN DCSP

### 2.1. CSP and Distributed CSP

**Definition 1 (Constraint Satisfaction Problem (CSP)).**

*Given $n$ variables $x_1, x_2, ..., x_n$, whose values are taken from finite, discrete domains $D_1, D_2, ..., D_n$:*

$$x_i \in D_i \quad \text{where } i = 1, 2, \ldots, n. \tag{1}$$

*and a set of constraints on their values, where a constraint is defined by a predicate:*

$$p_k(x_{k_1}, x_{k_2}, \ldots, x_{k_j}) \in D_{k_1} \times \cdots \times D_{k_j} \tag{2}$$

*the predicate is true if and only if the value assignment of these variables satisfies the constraint.*

*Solving a CSP is equivalent to finding an assignment of values for all variables such that all constraints are satisfied.*

In a Constraint Satisfaction Problem (CSP) the goal is to find a consistent assignment of values for a set of variables [1]. In general, there is no restriction about the form of the predicate for constraint. It can be a logical or mathematical formula, or any arbitrary relation defined by a tuple of variable values. We will sometimes also refer to these constraints as nogoods.

**Definition 2 (Distributed Constraint Satisfaction Problem (DCSP)).**

*A **distributed CSP** is a CSP in which the variables and constraints are distributed among automated agents.*

[4–6] Solving a distributed CSP can be considered as achieving coherence among the agents. Many application problems in DAI, such as interpretation problems, assignment problems, and multiagent truth maintenance tasks, can be formalized as distributed CSP.

For the agents, it assumes the following communication model:

- Agents communicate by sending messages. An agent can send messages to other agents if and only if the agent knows the addresses of the agents.

- The delay in delivering a message is finite, though random. For the transmission between any pair of agents, messages are received in the order in which they were sent.

In this model, the physical communication network may not be fully connected. In other words, the topology of the physical communication network does not play an important role here, it assumes the existence of a reliable underlying communication structure among the agents and ignores the implementation of the physical communication network.

Every agent owns some variables and it tries to determine their values. However, there exist inter-agent constraints which must be satisfied. Formally, there exist m agents *1, 2, ..., m*. Each variable $x_j$ belongs to one agent $i$, which could be represented as belongs($x_j$, *i*). Constraints are also distributed among agents. The fact that an agent *l* knows a constraint predicate $p_k$ is represented as known($p_k$, *l*).

A Distributed CSP is solved if and only if the following conditions are satisfied: for any $i$ and $x_j$ where $belongs(x_j, i)$, the value of $x_j$ is $d_j$, and for any $l$ and $p_k$ where $known(p_k, l)$, $p_k$ is true under the assignment $x_j = d_j$.

### 2.2. Hyper-Resolution-Based Consistency Algorithm and Polynomial Space Problem

Hyper-resolution-based consistency algorithm is a basic algorithm used to solve CSP problem. Some other famous algorithms such as Asynchronous Backtracking algorithm [6] and Asynchronous Weak-Commitment Search algorithm [9] can be considered as an extension of this algorithm. The core of this hyper-resolution-based consistency algorithm is using hyper-resolution rule, which is described as follows [1]:

**Theorem 1 (Hyper-Resolution Rule).**

$$\frac{\begin{array}{c} A_1 \vee A_2 \vee A_3 \cdots A_m \\ \neg(A_1 \wedge A_{11} \cdots) \\ \neg(A_2 \wedge A_{21} \cdots) \\ \vdots \\ \neg(A_m \wedge A_{m1} \cdots) \end{array}}{\neg(A_{11} \wedge \cdots \wedge A_{21} \wedge \cdots \wedge A_{m1} \cdots)} \quad (3)$$

*Where $A_i$ is a proposition.*

We can use hyper-resolution rule to solve DCSP problem. First, we map the elements of DCSP problem to above hyper-resolution rule, then the domain information of one variable can be represented by the first row of equation (3), where each proposition $A_i$ represents a possible domain value. Initial constraints related to this variable can be represented by the nogoods as in the next rows of equation (3). For each variable, we could represent the domain information and constraints similarly, and then use above hyper-resolution rule to generate new nogoods. These new nogoods could be communicated to related variables. Once the related variables get the new nogoods, they can update their constraint database, and use above rule to generate new nogoods based on the communicated nogood. Continuing above operations, if an empty nogood is generated, then the problem is over-constraint and has no solution. The hyper-resolution- based consistency algorithm is also described as in [1].

For example, for the graph coloring problem with 3 nodes connected to each other, and with color limited to 2 kinds, the table 1 lists domain value and initial constraints for each variable, where we use value 1 and 2 to represent the 2 different colors.

We assume $x_1$ generates a new nogood $\neg(x_2 = 1 \quad x_3 = 2)$ using nogood $\neg(x_1 = 1 \quad x_2 = 1)$ and nogood $\neg(x_1 = 2 \quad x_3 = 2)$ by hyper-resolution rule. This nogood is communicated to $x_2$ and $x_3$. $x_2$ generates a new nogood $\neg(x_3 = 2)$ using this communicated nogood and nogood $\neg(x_2 = 2 \quad x_3 = 2)$. Similarly, $x_1$ generates a new nogood $\neg(x_2 = 2 \quad x_3 = 1)$ from nogood $\neg(x_1 = 2 \quad x_2 = 2)$ and nogood $\neg(x_1 = 1 \quad x_3 = 1)$. $x_2$ generates a new nogood $\neg(x_3 = 1)$ using this nogood and nogood $\neg(x_2 = 1 \quad x_3 = 1)$. Then $x_3$ generates an empty nogood from nogood $\neg(x_3 = 2)$ and nogood $\neg(x_3 = 1)$. In other words, an empty nogood means nothing is good, which represents that the problem is over-constrained and has no solution. In this case, it is easy to understand for the last usage of the hyper-resolution rule – the domain information for $x_3$ says $x_3$ could be 1 or 2, however one nogood says $x_3$ can not be 1, another says $x_3$ can not be 2, so that it is not possible to find a value for $x_3$, which means no solution.

|  | $x_1$ | $x_2$ | $x_3$ |
|---|---|---|---|
|  | $(x_1 = 1 \vee x_1 = 2)$ | $(x_2 = 1 \vee x_2 = 2)$ | $(x_3 = 1 \vee x_3 = 2)$ |
|  | $\neg(x_1 = 1 \wedge x_2 = 1)$ | $\neg(x_2 = 1 \wedge x_1 = 1)$ | $\neg(x_3 = 1 \wedge x_1 = 1)$ |
|  | $\neg(x_1 = 1 \wedge x_3 = 1)$ | $\neg(x_2 = 1 \wedge x_3 = 1)$ | $\neg(x_3 = 1 \wedge x_2 = 1)$ |
|  | $\neg(x_1 = 2 \wedge x_2 = 2)$ | $\neg(x_2 = 2 \wedge x_1 = 2)$ | $\neg(x_3 = 2 \wedge x_1 = 2)$ |
|  | $\neg(x_1 = 2 \wedge x_3 = 2)$ | $\neg(x_2 = 2 \wedge x_3 = 2)$ | $\neg(x_3 = 2 \wedge x_2 = 2)$ |

Table 1. Constraints for example.

Above example gives us a clear idea about how the algorithm solves the DCSP problem. However, above example starts from variable *x₁* only and ignores other possible nogoods. Practically, if we use 3 agents work for each variable, and work asynchronously, we can generate a very large number of nogoods, which is in polynomial. In this case, for each variable, such as for *x₁* , we can choose any one of the two nogoods with *x₁* = *1*, and any one of two nogoods with *x₁* = *2* to generate new nogoods, so that we can generate 4 nogoods from the initial nogood set. For 3 variables, we can then generate 12 nogoods totally. Once these 12 nogoods communicate to each other, we can imagine how the knowledge base is increased.

## 3. HYPER-RESOLUTION-BASED CONSISTENCY ALGORITHM WITH E CIENT KNOWLEDGE BASE MANAGEMENT

### 3.1. Theorems Derived from Hyper-Resolution Rule

Before describing our management idea, we give several definitions and theorems as follows.

**Definition 3 (Sufficient Constraint).**
 Let nogood or constraint be:

$$b = \neg(a_1 \wedge a_2 \cdots \wedge a_m)$$
$$b' = \neg(a_1^* \wedge a_2^* \cdots \wedge a_n^*) \qquad (4)$$

where $m \leq n$ and set $\{a_k \mid \forall k \in \{1, 2, \cdots, m\}\} \subseteq \{a_k^* \mid \forall k \in \{1, 2, \cdots, n\}\}$, then we call that nogood $b$ is a **Sufficient Constraint** of *b'*.

**Theorem 2 (Nogoods Generation).**
 Assume domain information $A = A_1 \vee A_2 \vee A_3 \cdots A_m$, and let set $A* = \{A_i \mid i = 1, 2, \cdots, m, \text{ and } m \in Z^+\}$, where $Z^+$ donates the set of positive integers; nogoods set $B = \{\neg(A_i \wedge A_{ij1} \wedge A_{ij2} \cdots \wedge A_{ijk}) \mid A_i \in A*, A_{ij1} \wedge A_{ij2} \cdots \wedge A_{ijk} \in B_i, i = 1, 2, \cdots, m \text{ and } j, k \in Z^+\}$, where set $B_i$ is the collection of the rest part of the nogoods by removing $A_i$. Specially, if $\neg A_i \in B$, then we have $\epsilon \in B_i$, where $\epsilon$ donates empty symbol, which means the rest part is empty.
 If $\forall i \in \{1, 2, \cdots, m\}$, we have $B_i$ not empty, then we can generate new nogoods by using Hyper-Resolution Rule, and get new nogoods set:

$$C = \{\neg(b_1 \wedge b_2 \cdots \wedge b_m) \mid \forall b_i \in B_i, i = 1, 2, \cdots, m\} \qquad (5)$$

For example, assume *x₁* has domain information *A = (x₁ = 1   x₁ = 2)*. Correspondingly, *A = {A₁ , A₂ }*, where *A₁ = (x₁ = 1)* and *A₂ = (x₁ = 2)*. Assume *x₁* has nogoods set *B = {¬(x₁ = 1   x₃ = 1), ¬(x₁ = 1   x₃ = 2), ¬(x₁ = 2   x₂ = 2)}*, then we have *B₁ = {x₃ = 1, x₃ = 2}* and *B₂ = {x₂ = 2}*. By using above theorem 2, we can get new nogoods set *C = {(x₃ = 1 x₂ = 2), (x₃ = 2   x₂ = 2)}*.

**Theorem 3.** *Nogood set generated by theorem* **Nogoods Generation** *is complete.*

*Proof.* The theorem 2 generates all possible nogoods by hyper-resolution rule. Since the new nogoods set is the collection of all the nogoods generated, obviously it is complete.

**Definition 4 (False Nogood / False Constraint).**
   Let nogood or constraint be:

$$B = \neg(b_1 \wedge b_2 \cdots \wedge b_n) \qquad (6)$$

   *If there are some predicate $b_i$ be always false, where $i \in 1, 2, \cdots, n$, we call B a* **False Nogood** *or* **False Constraint**.

**Lemma 1.** *Let nogood or constraint be:*

$$B = \neg(b_1 \wedge b_2 \cdots \wedge b_n) \qquad (7)$$

   *If there are some predicates $b_i$ and $b_j$ that can not be satisfied at the same time, where $i, j \in 1, 2, \cdots, n$, then B is a False Nogood.*

*Proof.* When predicates $b_i$ and $b_j$ that can not be satisfied at the same time, $b_i \wedge b_j$ will be always false, and it can be looked as a predicate also. According to the definition 4, since $b_i \wedge b_j$ is always false, we have B to be a false nogood.

**Theorem 4 (Constraint Eliminating).** *In a nogoods set B, if a nogood $b_i$ satisfies one of following conditions:*

1. $\exists b_j \in B$ and $b_j \neq b_i$, such that $b_j$ is a sufficient constraint of $b_i$;
2. $b_i$ is a false nogood.

*Then $b_i$ can be eliminated from nogood set B without changing the completeness of B.*

*Proof.* Let's consider about the two conditions separately, if $b_j$ is a sufficient constraint of $b_i$, then according to the definition 3, we can write $b_j$ and $b_i$ to be the format:

$$b_j = \neg(a_1 \wedge a_2 \cdots \wedge a_m)$$
$$b_i = \neg(a_1^* \wedge a_2^* \cdots \wedge a_n^*)$$

where $m \leq n$ and set $\{a_k \mid k \in \{1, 2, \cdots, m\}\} \subseteq \{a_k^* \mid k \in \{1, 2, \cdots, n\}\}$. So that we can rewrite $b_i$ to be: $b_i = \neg(a_1 \wedge a_2 \cdots \wedge a_m \wedge a_k \cdots)$

If $b_j$ is satisfied, that is, when $b_j$ is true, then $a_1 \wedge a_2 \cdots \wedge a_m$ is false. Because $0 \wedge x = 0$, where 0 donates false and x donates any proposition, $a_1 \wedge a_2 \cdots \wedge a_m \wedge a_k \cdots$ will also be false, so $b_i$ is true. In other words, when $b_j$ is satisfied, $b_i$ will also be satisfied. So by removing $b_i$ from the nogood set B, we don't lose any constraint information.For the second condition, if bi is a false nogood, according to the definition 4, there is a proposition always false. Since $0 \quad x = 0, a_1 \quad a_2 \cdots \quad a_n$ will be always false, so that bi will be always true. Since bi will be always satisfied, it does not supply any constraint information, so we can simply remove it from without change any thing.

### 3.2. Hyper-Resolution-Based Consistency Algorithm with Efficient Knowledge Base Management

Based on above derivations, we generate following algorithm:
    1. Initialize the knowledge base with domain information and nogoods;
    2. Generate new nogoods with false nogoods eliminated;
    3. Communicate new nogoods to related variables;
    4. Once the related variables receive the new nogoods, they update their knowledge base by remove all sufficient nogoods of the new added nogoods, and eliminate the new added nogoods if the new added nogoods are sufficient nogoods of some nogoods in the knowledge base.
    5. If the new nogood is not added, do nothing, otherwise generate new nogoods based on the new added nogoods.
    6. If empty nogoods is generated, report no solution and stop, otherwise, repeat 3 to 6.

Assume initial domain information is represented as $A = A_1 \quad A_2 \quad A_3 \cdots A_m$, and let set $A = \{A_i \mid i = 1, 2, \cdots, m, \text{ and } m \quad Z^+ \}$, where $Z^+$ donates the set of positive integers. Assume nogoods set $B = \{\neg(A_i \quad A_{ij\,1} \quad A_{ij\,2} \cdots \quad A_{ij\,k}) \mid A_i \quad A \}$, and we let $B_i = \{A_{ij\,1} \quad A_{ij\,2} \cdots \quad A_{ij\,k} \mid \neg(A_i \quad A_{ij\,1} \quad A_{ij\,2} \cdots \quad A_{ij\,k}) \quad B, A_i \quad A \text{ and } j, k \quad Z^+ \}$. Specially, if $\neg A_i \quad B$, then we have $B_i$, where donates empty symbol, which means the rest part without $A_i$ is empty.

For example, according to the initial domain information and nogoods as in Table 1, for $x_1$ we can get $A = \{x_1 = 1, x_1 = 2\}$, $B_1 = \{x_2 = 1, x_3 = 1\}$, $B_2 = \{x_2 = 2, x_3 = 2\}$.

Then the details of generate new nogoods could be described as follows:

```
NOGOODS-GENERATION(new-nogood)
1  if new-nogood is NULL,
      then generate new nogoods based on the whole knowledge base:
2         nogoods ← {¬(b₁ ∧ b₂ ··· ∧ bₘ) | ∀bᵢ ∈ Bᵢ, i = 1, 2, ···, m}
   else generate new nogood based on new-nogood
        Assume new-nogood = ¬(Aⱼ ∧ c₁ ∧ c₂ ···)
3         nogoods ← {¬(b₁ ∧ b₂ ··· ∧ bᵢ₋₁ ∧ (c₁ ∧ c₂ ···) ∧ bᵢ₊₁ ··· ∧ bₘ) |
                    ∀bᵢ ∈ Bᵢ, and i ≠ j}
4  for each nogood ∈ nogoods,
5      do if nogood is a false nogood,
6         then eliminate it from nogoods
```

**Fig. 1.** Nogoods Generation

Above procedure can be used for step 2 and 5. For step 4, instead of simply add the new nogoods in, based on above assumptions, the details of the procedure can be described as follows:

```
KNOWLEDGE-BASE-UPDATE(nogood-receive)
    Assume nogood-receive = ¬(A_j ∧ c_1 ∧ c_2 ···)
1   for each b_i ∈ B_j,
2       do compare ¬(A_j ∧ b_i) with nogood-receive
3           if ¬(A_j ∧ b_i) is a sufficient nogood of nogood-receive
4               then let nogood-receive be NULL
5                   Goto Label 1
6               else if nogood-receive is a sufficient nogood of ¬(A_j ∧ b_i)
7                   then remove ¬(A_j ∧ b_i) from B
8                       remove b_i from B_j
9   Label 1: if nogood-receive is not NULL
10      then add nogood-receive to B
11          add (c_1 ∧ c_2 ···) to B_j
```

Fig. 2. Knowledge Base Updating

By eliminating false nogoods in *NOGOODS–GENERATION* , we reduce the number of the nogoods to be communicated; by eliminating sufficient nogoods in *KNOWLEDGE–BASE–UP DATE*, we decrease the size of the knowledge base.

## 4. EXAMPLE

This section will give an example based on Table 1, and show how our approach decreases the knowledge base with comparison to the general one.

Example given in Section 2.2 just shows how the algorithm works to solve the problem, which focus on one variable, and ignore messages that are not used. Here, we show all messages produced and added to knowledge base, and compare the result with our approach.

Following table 2 shows the knowledge base changes based on the general hyper-resolution-based consistency algorithm. In this table the constraints are proceeded by a number which indicated the time at which they were added.

By comparing above two approaches, we see that each variable initially has 4 nogoods, based on which each can generate 4 new nogoods, because we have 2 elements in $B_1$ and 2 in $B_2$ . However in the second approach, by removing false nogoods, such as ¬($x_2 = 1$    $x_2 = 2$) and ¬($x_3 = 1$    $x_3 = 2$) for $x_1$ , we get only 2 new nogoods for each variable to communicate. After first communication, there are 5 nogoods added to each variable for the first approach, and 4 added to each variable for the second approach, as in Table 2 and 3.

Then based on the new nogoods, the first approach can generate 21 new nogoods for each variable. For example, for variable $x_1$ , since there are 3 with $x_1 = 1$ and 3 with $x_2 = 2$ in the new received nogood, then they could generate 3 × 2 + 3 × 2 nogoods with old nogoods in the knowledge base, and they could generate 3 × 3 nogoods among the new nogoods. By communicating these nogoods to related variables, there are only 2 that can be really added to the knowledge base for each variable, since most of them are redundant (already in the knowledge base). For the second approach, based on the 4 new received nogoods, each variable can generate 10 new nogoods. For example, for $x_1$ , since there are 2 with $x_1 = 1$ and 2 with $x_2 = 2$ in the new received nogood, then they could generate 2 × 2 + 2 × 2 nogoods with old nogoods in the knowledge base, and they could generate 2 × 2 nogoods among the new nogoods, and by removing false nogoods ¬($x_2 = 1$    $x_2 = 2$) and ¬($x_3 = 1$    $x_3 = 2$), we have 10 new nogoods left. By communicating these nogoods to related variables, there are only 2 added to knowledge base, with same reason in the first approach. However, when these 2 new nogoods are added, which are sufficient nogoods for some nogoods in the knowledge base, so that those nogoods are eliminated. For example, for $x_1$, once $\neg(x_1 = 1)$ is added, which is sufficient nogood of all other nogoods with $x_1 = 1$, so that all such nogoods with $x_1 = 1$ such as 01, 02, 11 and 13 are eliminated. Similarly, once $\neg(x_1 = 2)$ is added, 03, 04, 12, and 14 can be eliminated.

| $x_1$ | $x_2$ | $x_3$ |
|---|---|---|
| $0 : (x_1 = 1 \vee x_1 = 2)$ | $0 : (x_2 = 1 \vee x_2 = 2)$ | $0 : (x_3 = 1 \vee x_3 = 2)$ |
| $0 : \neg(x_1 = 1 \wedge x_2 = 1)$ | $0 : \neg(x_2 = 1 \wedge x_1 = 1)$ | $0 : \neg(x_3 = 1 \wedge x_1 = 1)$ |
| $0 : \neg(x_1 = 1 \wedge x_3 = 1)$ | $0 : \neg(x_2 = 1 \wedge x_3 = 1)$ | $0 : \neg(x_3 = 1 \wedge x_2 = 1)$ |
| $0 : \neg(x_1 = 2 \wedge x_2 = 2)$ | $0 : \neg(x_2 = 2 \wedge x_1 = 2)$ | $0 : \neg(x_3 = 2 \wedge x_1 = 2)$ |
| $0 : \neg(x_1 = 2 \wedge x_3 = 2)$ | $0 : \neg(x_2 = 2 \wedge x_3 = 2)$ | $0 : \neg(x_3 = 2 \wedge x_2 = 2)$ |
| (Generate 4 nogoods) | (Generate 4 nogoods) | (Generate 4 nogoods) |
| $1 : \neg(x_1 = 1 \wedge x_3 = 2)$ | $1 : \neg(x_2 = 1 \wedge x_3 = 2)$ | $1 : \neg(x_3 = 1 \wedge x_2 = 2)$ |
| $1 : \neg(x_1 = 1 \wedge x_1 = 2)$ | $1 : \neg(x_2 = 1 \wedge x_2 = 2)$ | $1 : \neg(x_3 = 1 \wedge x_3 = 2)$ |
| $1 : \neg(x_1 = 2 \wedge x_3 = 1)$ | $1 : \neg(x_2 = 2 \wedge x_3 = 1)$ | $1 : \neg(x_3 = 2 \wedge x_2 = 1)$ |
| $1 : \neg(x_1 = 1 \wedge x_2 = 2)$ | $1 : \neg(x_2 = 1 \wedge x_1 = 2)$ | $1 : \neg(x_3 = 1 \wedge x_1 = 2)$ |
| $1 : \neg(x_1 = 2 \wedge x_2 = 1)$ | $1 : \neg(x_2 = 2 \wedge x_1 = 1)$ | $1 : \neg(x_3 = 2 \wedge x_1 = 1)$ |
| (Generate 21 nogoods) | (Generate 21 nogoods) | (Generate 21 nogoods) |
| $0 : \neg(x_1 = 1)$ | $0 : \neg(x_2 = 2)$ | $2 : \neg(x_3 = 2)$ |
| $0 : \neg(x_1 = 2)$ | $0 : \neg(x_2 = 1)$ | $2 : \neg(x_3 = 1)$ |
| (Generate 11 nogoods with an empty nogood) | (Generate 11 nogoods with an empty nogood) | (Generate 11 nogoods with an empty nogood) |

Table 2. Example for general version

| $x_1$ | $x_2$ | $x_3$ |
|---|---|---|
| $0 : (x_1 = 1 \vee x_1 = 2)$ | $0 : (x_2 = 1 \vee x_2 = 2)$ | $0 : (x_3 = 1 \vee x_3 = 2)$ |
| $01 : \neg(x_1 = 1 \wedge x_2 = 1)$ | $01 : \neg(x_2 = 1 \wedge x_1 = 1)$ | $01 : \neg(x_3 = 1 \wedge x_1 = 1)$ |
| $02 : \neg(x_1 = 1 \wedge x_3 = 1)$ | $02 : \neg(x_2 = 1 \wedge x_3 = 1)$ | $02 : \neg(x_3 = 1 \wedge x_2 = 1)$ |
| $03 : \neg(x_1 = 2 \wedge x_2 = 2)$ | $03 : \neg(x_2 = 2 \wedge x_1 = 2)$ | $03 : \neg(x_3 = 2 \wedge x_1 = 2)$ |
| $04 : \neg(x_1 = 2 \wedge x_3 = 2)$ | $04 : \neg(x_2 = 2 \wedge x_3 = 2)$ | $04 : \neg(x_3 = 2 \wedge x_2 = 2)$ |
| (Generate 2 nogoods) | (Generate 2 nogoods) | (Generate 2 nogoods) |
| $11 : \neg(x_1 = 1 \wedge x_3 = 2)$ | $11 : \neg(x_2 = 1 \wedge x_3 = 2)$ | $11 : \neg(x_3 = 1 \wedge x_2 = 2)$ |
| $12 : \neg(x_1 = 2 \wedge x_3 = 1)$ | $12 : \neg(x_2 = 2 \wedge x_3 = 1)$ | $12 : \neg(x_3 = 2 \wedge x_2 = 1)$ |
| $13 : \neg(x_1 = 1 \wedge x_2 = 2)$ | $13 : \neg(x_2 = 1 \wedge x_1 = 2)$ | $13 : \neg(x_3 = 1 \wedge x_1 = 2)$ |
| $14 : \neg(x_1 = 2 \wedge x_2 = 1)$ | $14 : \neg(x_2 = 2 \wedge x_1 = 1)$ | $14 : \neg(x_3 = 2 \wedge x_1 = 1)$ |
| (Generate 10 nogoods) | (Generate 10 nogoods) | (Generate 10 nogoods) |
| $2 : \neg(x_1 = 1)$ | $2 : \neg(x_2 = 2)$ | $2 : \neg(x_3 = 2)$ |
| (Eliminate 01,02,11,13) | (Eliminate 03,04,12,14) | (Eliminate 03,04,12,14) |
| $2 : \neg(x_1 = 2)$ | $2 : \neg(x_2 = 1)$ | $2 : \neg(x_3 = 1)$ |
| (Eliminate 03,04,12,14) | (Eliminate 01,02,11,13) | (Eliminate 01,02,11,13) |
| (Generate 1 empty nogood) | (Generate 1 empty nogood) | (Generate 1 empty nogood) |

Table 3. Example for the version with efficient knowledge base management

So far, at step 2, there are only 2 nogoods left for each variable in the knowledge base for the second approach, while there are 4 + 5 + 2 nogoods left for each variable for the first approach. Thus, in the last step, the first approach generate 11 nogoods (1 × 5 + 1 × 5 + 1 × 1), among them, there is an empty nogood, which terminate the program by report no solution. For the second approach, we only generate one empty nogood with the left nogood for each variable.

International Journal of Ad hoc, Sensor & Ubiquitous Computing (IJASUC) Vol.1, No.3, September 2010

## 5. EVALUATION

From above example in Section 4, we can see that how our approach reduces the new nogoods generated for communication, and how our approach decreases the knowledge base when possible. While the general hyper-resolution-based approach never decreases the knowledge base. The efficiency of our approach is obvious. The comparison of the two algorithms: the original Hyper-resolution-based consistency algorithm and the one with Efficient Knowledge Base Management (EKBM), based on example is given in Figure 3.

From the Figure, we can find out that the algorithm with EKBM generates less nogoods for communication than the original algorithm. And the knowledge space used for storing the nogoods keeps increasing in the original algorithm, there is no deceasing process. However, in the algorithm with EKBM, it involves a decreasing process. Thus, the provided EKMB method helps to maintain a minimal required knowledge space. This will also speed up the searching process by reducing the search space.

Other than the efficiency showed in above example, the correctness and completeness for the EKBM approach are proved in Section 3.1 as well.

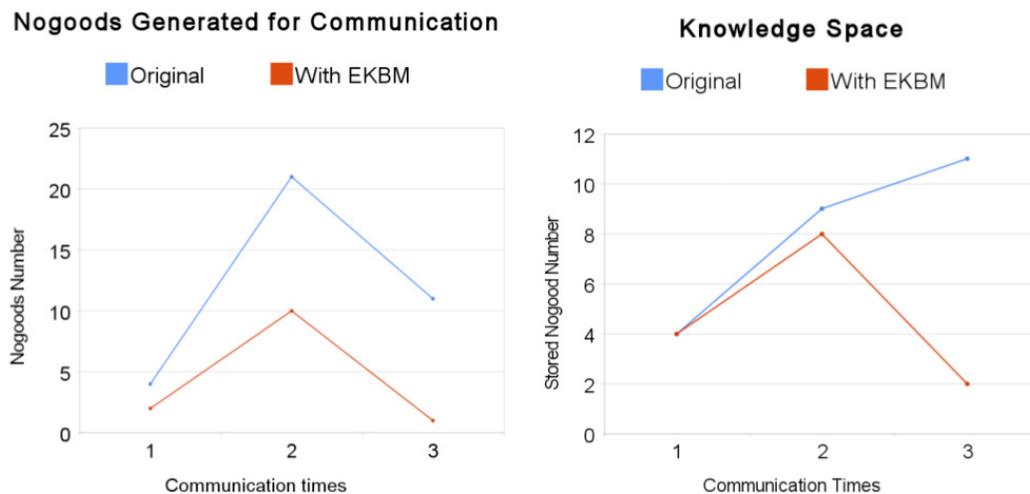

Fig. 3. Comparison of the Two Algorithms on the Given Example

The Asynchronous Backtracking (ABT) algorithm and asynchronous weak-commitment search can be considered as an extension of hyper-resolution based consistency algorithm. The difference between ABT and hyper-resolution based consistency algorithm is that the former communicates the new generated no-goods to higher priority variable only, while the later sends the nogoods to all related variable; asynchronous weak-commitment search algorithm also considers the priority in, however not like ABT in which the priority of variables are determined, asynchronous weak-commitment search can change the priority of the variables. Disregarding the priority of the variables, we can directly apply above *NOGOODS− GENERATION* and *KNOWLEDGE − BASE − UPDATE* into ABT and asynchronous weak-commitment search, which benefits the implement the same way as above.

## 6. CONCLUSION

This paper provides an efficient knowledge base management approach based on general usage of hyper-resolution-rule in consistence algorithm. The approach minimizes the increasing of the knowledge base by eliminating sufficient constraint and false nogood. These eliminations do not change the correctness and completeness of the original knowledge base increased. The proofs are given as well.

The given example shows how this new approach with EKMB reduces the new nogoods generated for communication, and how it decreases the knowledge base during the searching process. The comparison of the original algorithm and the new algorithm is given as well. It shows that the algorithm with EKBM generates less nogoods for communication than the original algorithm, and it involves a decreasing process. However the original hyper-resolution-based approach never decreases the knowledge base.

Thus, the provided EKMB method helps to maintain a minimal required knowledge space, and will potentially speeds up the searching process by reducing the search space and simplifying the searching process. With the problem size increasing, the benefit of this approach will be more obvious.

Considering that the provided EKBM approach is on a general base, all other algorithms adopting hyper-resolution-rule can take the advantage of this approach, such as the Asynchronous Backtracking (ABT) algorithm and asynchronous weak- commitment search algorithm in solving DCSP.